# Robustness Evaluation of GPU Cloud Pricing Benchmarks under Incomplete Observations: A Controlled Missingness Study

Jooya Wong
Chongqing College of Mobile Communication, Chongqing, China
jooyawong@cqytxy.edu.cn

***Abstract*—Daily GPU rental-price benchmarks can fail when marketplace quotes are missing. This article stress-tests a fixed equal-weight median rule under four imposed masking mechanisms and four benchmark-level recovery methods. Across 5,800 Monte Carlo realisations in 27 scenarios on 190 held-out days, whole-source outages reduce publication coverage to 0.77 while preserving zero conditional error on published dates; composition-changing loss retains 0.86 coverage but produces 7.74 log points of error. A separate past-only stratum-level ablation reduces error by 4.15–6.70 log points in five composition-changing cases, using a richer information set. An observable-signal withholding rule improves the primary error–coverage criterion in 0 of 10 cases. Results concern observed asking-price measurements, not transaction prices or identified real-world missingness mechanisms.***

## I. INTRODUCTION

GPU rental marketplaces change continuously, while exporters can miss quotes or entire days. Daily price benchmarks support capacity and cost planning, making both publication availability and the accuracy of published levels operational concerns. Archived logs rarely identify the cause of a gap; this article studies the effects of incomplete observation instead.

Cloud monitoring pipelines aggregate heterogeneous rental offers into daily model-level prices [11], [12], with quality and observability constraints [16], [17]. Loss of individual strata may bias the aggregate even when a value can be published, whereas a full-feed outage may prevent publication. These outcomes require joint evaluation of error and coverage.

Unlike index-composition studies that alter weights or baskets [5], [8], this article fixes the measurement rule and varies the observation process. The objective is reliability of a cloud-data measurement, not prediction of transaction prices.

Three layers are distinguished: observation loss (visible quotes), measurement error (deviation from the full-observation benchmark), and publication reliability (whether an estimate is available). Whole-date outages mainly affect availability; structural stratum loss mainly affects level accuracy.

RQ1 asks how random, clustered and group-dependent losses affect error and coverage. RQ2 examines recovery-method rankings across mechanisms, rates and interruption lengths on common published dates. RQ3 asks whether observable coverage, staleness and stratum signals justify withholding. H1–H3 correspond to mechanism-specific effects, ranking changes and coverage-aware withholding.

Contributions are: (i) an information-isolated, reproducible missingness stress test with complete-date outage semantics; (ii) joint and paired evaluation of four benchmark-level strategies; (iii) a separate past-only stratum-level reconstruction ablation (Table VII); and (iv) a withholding-rule test with thresholds selected before independent evaluation.

The feed contains posted asking prices, not transactions; its GPU-count field is a supply gauge, strata are not providers, and missing records do not establish absent supply [10]. The reference is an observed-data benchmark under one fixed aggregation rule.

## II. RELATED WORK

### A. Missing-data mechanisms and selection-induced bias

Rubin [1] and Little and Rubin [2] formalise missing-data mechanisms; Dempster et al. [6] develop likelihood treatment, and Little [7] addresses MCAR testing. Schafer and Graham [3] discuss complete-case and single-imputation bias. Here MCAR-, MAR- and MNAR-like labels describe imposed masks, not the unknown exporter mechanism. Withholding also changes the dates used to calculate conditional error; common-date comparisons and coverage are therefore essential.

### B. Online versus offline reconstruction

Time-series imputation methods [4] and gap-reconstruction results for IoT and stationary series [18], [19] distinguish available information from the recovery algorithm. Past-only carrying and rolling estimation are usable online; interpolation requiring future observations is retrospective. Circular block-bootstrap bands [9] describe serial variation within the test window, not significance of cross-mechanism contrasts.

### C. Price measurement, observability and availability

The CPI Manual [5] supplies context on missing quotations; cloud spot-price measurement [8], [11]–[14] concerns related but different markets. Benchmark sensitivity [15] and data-quality frameworks [16], [17] motivate reporting completeness, freshness and accuracy together. No cited source supplies ground truth for unobserved GPU rental offers.

## III. DATA AND REFERENCE MEASUREMENT

### A. Audit and asset classes

Three asset classes were audited: (A) Vast.ai / 500.farm grouped daily exporter history [10], 756,152 rows over 2022-10-01–2026-11-06; (B) mostly undated multi-provider quotation snapshots; and (C) weekly or monthly aggregates. The master extends an older frozen copy of the same feed, not an independent dataset. Only Class A is analysed read-only.

### *B. Panel and the fixed reference rule*

A cell denotes one marketplace stratum (datacenter flag × machine-size band), at most six per GPU. Valid quotes have positive price and gauge and a normal quality flag. The 381-day window (2025-02-22–2026-03-09) covers five models, 9,928 valid quotes and chronological training/validation/test periods of 99/92/190 days. Seven RTX 5090 training GPU-dates lack a defined reference.

The fixed reference is the equal-weight median of at least three valid stratum quotes per GPU-day. Errors use $100 \times |\ln(\text{estimate}/\text{reference})|$ log points; small values approximate percentage errors but are not percentages. The alternative gauge-weighted-mean sensitivity check changes both estimate and reference (Section V-E). All test-period GPU-days have defined references.

TABLE I. DATA AND SAMPLE

| Item | Value |
|---|---|
| Primary source | Vast.ai / 500.farm daily exporter [10] |
| Full file span | 2022-10-01..2026-11-06; 756,152 rows |
| Study window | 2025-02-22 .. 2026-03-09 (381 d) |
| Train / Val / Test | 99 / 92 / 190 d |
| GPU models | H100 SXM, A100 SXM4, 3090, 4090, 5090 |
| Cells per GPU | 6 marketplace strata (not providers) |
| Valid quotes | 9,928 in window |
| Reference undefined | 7 (GPU, date); 0 in TEST |
| Price / gauge | USD/GPU-h asking p50 / mean GPU gauge |

### *C. Real missingness versus artificial masks*

Real gaps include seven days with no feed rows, a six-day model-specific partial outage and seven under-observed RTX 5090 training dates. The study window excludes the outage episodes but retains the seven training dates without reference values. Artificial masks affect observed valid cells only; actual missing values are never treated as ground truth or imputed into the reference.

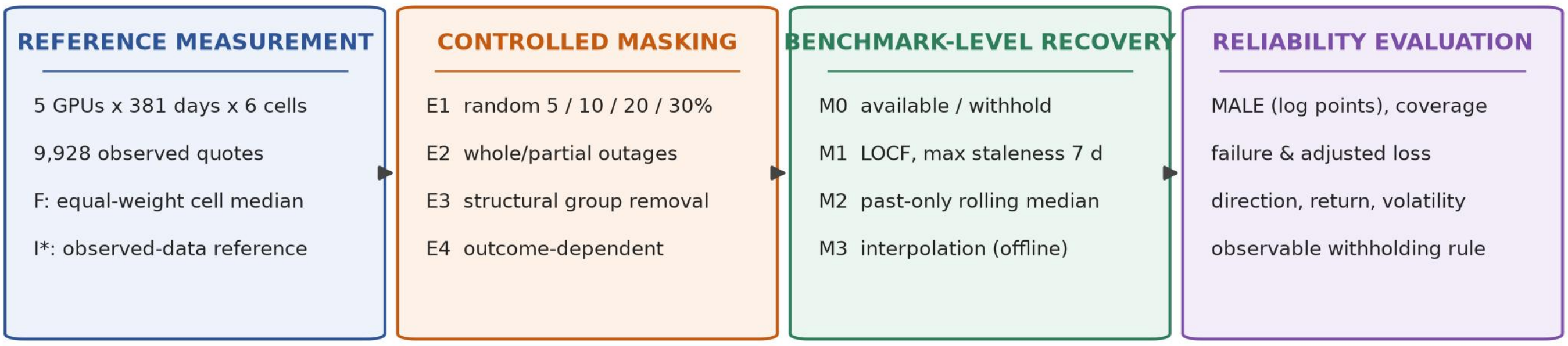

Fig. 1. Controlled-missingness workflow and recovery levels.

## IV. CONTROLLED MISSINGNESS METHODOLOGY

### *A. Three evaluation layers*

Observation loss is the masked-cell rate and number of complete dark days; measurement error is deviation of published levels; reliability is publication coverage and its complement, failure rate. Withholding can reduce conditional error while worsening availability, so these measures are assessed jointly.

### *B. Four masking mechanisms*

Each E1 random mask drops valid cells independently at 5%, 10%, 20% or 30%. E2 imposes whole-source or half-cell partial outages lasting 1, 3, 7 or 14 consecutive days, separated by clear days; complete dark dates are never trimmed to match nominal loss. E3 preferentially masks strata by training-period price, gauge or 8+ configuration. E4 depends on the observed quote, daily price change or inverse gauge (MNAR-like). These are experimental interventions, not inferred collector causes.

TABLE II. MASKING SCENARIOS

| ID | Mechanism | Parameters | Repeats |
|---|---|---|---|
| E1 | Random (MCAR-like) | q = 5 / 10 / 20 / 30% | 500 |
| E2 | Complete-date outages | L = 1/3/7/14 d; W / P | 200 |
| E3 | Structural removal | price, gauge, 8+ config | 200 |
| E4 | Outcome-dependent | quote, change, gauge | 200 |

E1: 500 repeats; E2–E4: 200. Whole-day loss is realised, not forced.

### *C. Observable publication gate*

The publication gate uses only visible valid-cell count, visible gauge and a past-only gauge expectation. The pre-mask full-gauge ratio is an offline diagnostic and never enters publication decisions.

The gauge expectation uses the preceding $W_g = 28$ positive-gauge days; publication requires at least three visible cells and a gauge ratio ≥ 0.60. M2 uses $W_{M2} = 7$ days and M1/M4 use maximum staleness $S_{max} = 7$ days. Supplementary Test A perturbs hidden prices, gauges and the oracle denominator: observable decisions and online estimates remain invariant while the oracle changes.

### *D. Benchmark-level recovery and withholding*

On each common mask, M0 publishes only when the gate passes; M1 carries the last published level for at most seven days; M2 uses the past seven-day median of published levels; M3 interpolates between bracketing published levels and is offline. These four methods reconstruct aggregates, not individual strata. Withheld dates count as failures, never zero-error publications.

The separate M4 ablation carries each stratum's past quote forward for at most seven days and re-aggregates under the same rule. M4 and M1 share the publication gate and benchmark fallback; their paired contrast isolates stratum-level recovery but gives M4 a richer information set, precluding equal-information algorithmic claims.

Withholding thresholds use observable active strata, staleness, consecutive gaps and gauge ratio, selected on training and validation data and frozen for the test period.

The primary withholding criterion is $J(\lambda) = \text{conditional error} + 100\lambda(1 - \text{coverage})$, with $\lambda = 0.05$ and sensitivity at 0.01 and 0.10. A distinct coverage-adjusted loss averages published absolute log errors plus a 10-log-point penalty per withheld date, also tested at penalties 5 and 20. The two criteria have different normalisers; failure costs are stipulated, not externally calibrated.

### *E. Metrics and inference*

Conditional MALE measures absolute log-point deviation on published dates; Î denotes the estimate and I* the observed-data reference. MAPE and trend metrics are secondary.

Coverage is the proportion of dates published; failure rate is its complement. Directional disagreement, return MAE and volatility ratio test trend fidelity. Paired MALE is also computed on dates both methods publish and on the four-method intersection; adjusted loss penalises unavailable outputs.

E1 uses 500 Monte Carlo masks per scenario and E2–E4 use 200, with recorded seeds. Repetitions resample missingness on a single historical window, not independent markets. The reported 5–95% ranges describe mask-draw dispersion; seven-day circular block bootstraps describe serial dependence of published-date errors. Neither supports cross-mechanism significance claims.

## V. EXPERIMENTAL RESULTS

### *A. Random loss rate*

Under random loss, M1 conditional MALE rises from 1.96 at 5% masking to 6.81 at 30%; M0 rises from 1.63 to 6.35 (Fig. 2a). At 20%, M0 coverage is only 0.78, giving adjusted loss 6.08 against M2's 4.67. Across 500 draws, M0 conditional MALE spans 4.62–5.42 (5th–95th percentiles).

### *B. Mechanism-specific error and availability*

At nominal 20% loss, whole-source seven-day outages leave M0's published values exact (conditional MALE 0.00) but lower its coverage to 0.77. Low-price stratum removal preserves 0.86 coverage yet induces 7.74 log points of error (Fig. 2b; Table III). Thus source outages primarily threaten publication, whereas composition-changing loss threatens measurement accuracy; the zero error under whole-date masking is algebraic, not recovery success.

**TABLE III. CONDITIONAL MALE AND M0 COVERAGE**

| Scenario | M0 | M1 | M2 | M3 | M0 cov. |
|---|---|---|---|---|---|
| Random 20% | 4.99 | 5.40 | 4.66 | 5.12 | 0.78 |
| Outage whole 7d | 0.00 | 0.94 | 2.89 | 0.80 | 0.77 |
| Outage partial 7d | 2.14 | 2.86 | 4.32 | 2.68 | 0.72 |
| Struct. high-price | 5.53 | 5.79 | 6.72 | 5.62 | 0.82 |
| Struct. low-price | 7.74 | 7.58 | 8.14 | 7.50 | 0.86 |
| Struct. 8+ config | 5.12 | 5.50 | 4.93 | 5.21 | 0.77 |
| MNAR-like high-price | 5.29 | 5.68 | 6.36 | 5.46 | 0.81 |
| MNAR-like low-gauge | 6.08 | 6.49 | 6.55 | 6.32 | 0.86 |

Means over 200 masks (E1: 500). M3 is offline; full metrics: S1.

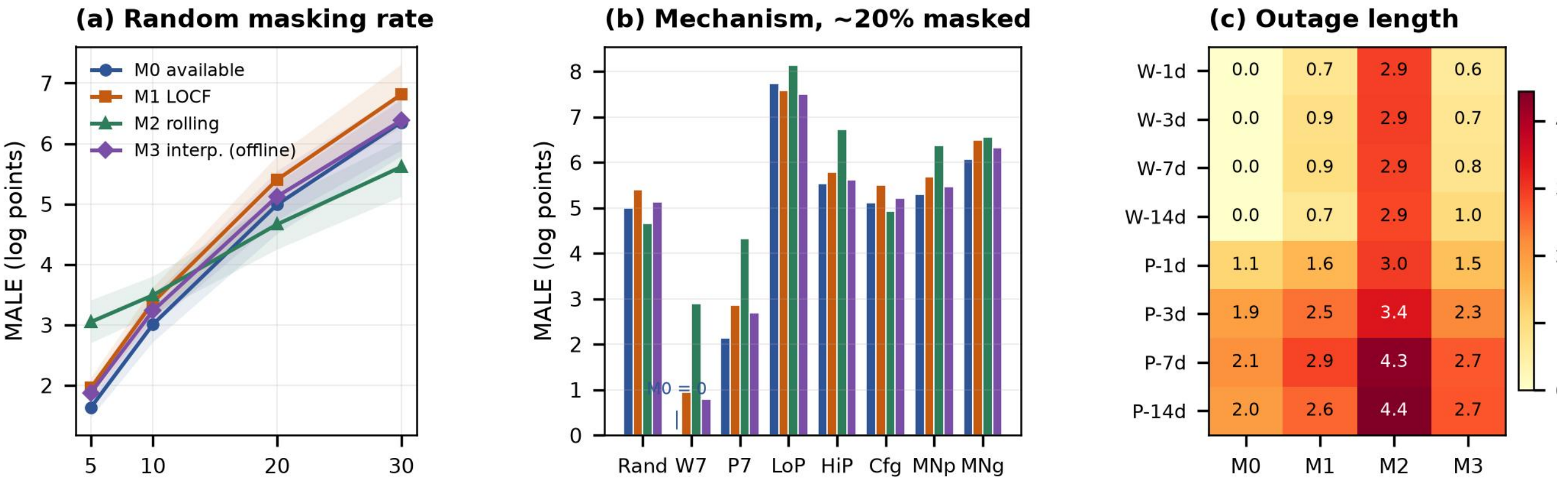

Fig. 2. Conditional MALE: (a) random loss; (b) mechanisms at ~20%; (c) outage length. W/P = whole/partial; LoP/HiP/Cfg = structural groups; MNp/MNg = MNAR-like groups.

Table IV varies whole-source outage length from 1 to 14 days. M0's published-date MALE remains zero while coverage ranges 0.73–0.77. With M1, L=7 to L=14 reduces conditional MALE from 0.94 to 0.71 but coverage from 0.99 to 0.88: the seven-day staleness cap excludes difficult dates. Unbounded carry-forward would yield 5.64 rather than 0.71 log points (S6). M3's error grows from 0.63 to 1.01; M2 stays near 2.90, with L=14 coverage 0.86.

**TABLE IV. OUTAGE COVERAGE AND MALE**

| L | Cov.0 | Cov.1 | Cov.2 | Cov.3 | M1 | M2 | M3 |
|---|---|---|---|---|---|---|---|
| 1 | 0.75 | 1.00 | 0.99 | 1.00 | 0.73 | 2.92 | 0.63 |
| 3 | 0.75 | 0.99 | 0.99 | 1.00 | 0.89 | 2.93 | 0.73 |
| 7 | 0.77 | 0.99 | 0.96 | 1.00 | 0.94 | 2.89 | 0.80 |
| 14 | 0.73 | 0.88 | 0.86 | 1.00 | 0.71 | 2.86 | 1.01 |

Cov.0–3 = publication coverage; M1–M3 = conditional MALE (log points). M0 = 0 by design; diagnostics: S6.

### *C. Joint reporting and common published dates*

At 20% random loss, Table V jointly reports conditional error, coverage, failure, adjusted loss, common-date error and trend metrics. M0's lower conditional MALE accompanies the lowest coverage; rankings change with common-date restriction or failure penalties. No universal recovery winner follows.

Different experimental masks explain M1's 5.40 in Table V versus 5.31 in the paired reference-rule check (Table VI); M4 uses another paired mask family (Table VII). Gauge-weighted mean gives M1 2.13 versus 5.31 log points under the median rule, so conclusions depend on the fixed aggregation rule.

**TABLE V. RANDOM 20% LOSS: JOINT METRICS**

| Method | MALE | Cov. | Fail. | Adj. | Com. | Dir. | Vol. |
|---|---|---|---|---|---|---|---|
| M0 | 4.99 | 0.78 | 0.22 | 6.08 | 4.99 | 0.35 | 2.84 |
| M1 | 5.40 | 1.00 | 0.00 | 5.41 | 4.99 | 0.50 | 2.43 |
| M2 | 4.66 | 1.00 | 0.00 | 4.67 | 4.31 | 0.60 | 0.82 |
| M3 (offline) | 5.12 | 1.00 | 0.00 | 5.13 | 4.99 | 0.38 | 2.28 |

MALE = conditional error; Cov./Fail. = coverage/failure; Adj. = penalised loss; Com. = common dates; Dir. = direction; Vol. = volatility. Details: S2–S3.

### D. Error-coverage trade-off and trend distortion

In Fig. 3a, conditional error is plotted against coverage; Fig. 3b tests withholding. At 20% random loss, M2 smooths the series (volatility ratio 0.82 versus M1's 2.43), but has 0.60 directional disagreement versus M0's 0.35. Low level error does not guarantee trend fidelity. M3 remains retrospective.

### E. Observable withholding and cost sensitivity

For primary λ = 0.05, validation selects maximum staleness 7 days, at least three active strata, gap ≤ 7 days and gauge ratio ≥ 0.3. Withholding reduces conditional MALE in all 10 cases (mean −0.25) but coverage also falls (mean −0.10), improving J in 0/10. Table VI shows the cost dependence, so default withholding lacks support.

**TABLE VI. WITHHOLDING COST AND RULE CHECK**

| Check | Result |
|---|---|
| H3 cases / favourable (primary) | 10 / 0 |
| Mean Δ MALE / Δ coverage | -0.25 / -0.10 |
| Cost λ = 0.01: favourable / mean ΔJ | 8 / 10; -0.41 |
| Cost λ = 0.05: favourable / mean ΔJ | 0 / 10; 0.25 |
| Cost λ = 0.1: favourable / mean ΔJ | 0 / 10; 0.75 |
| M1 MALE, paired F: median / gauge-wtd | 5.31 / 2.13 |
| Zero-mask MALE (both paired rules) | 0.00 |

Both reference rules match their own zero-mask baseline; comparison uses paired masks.

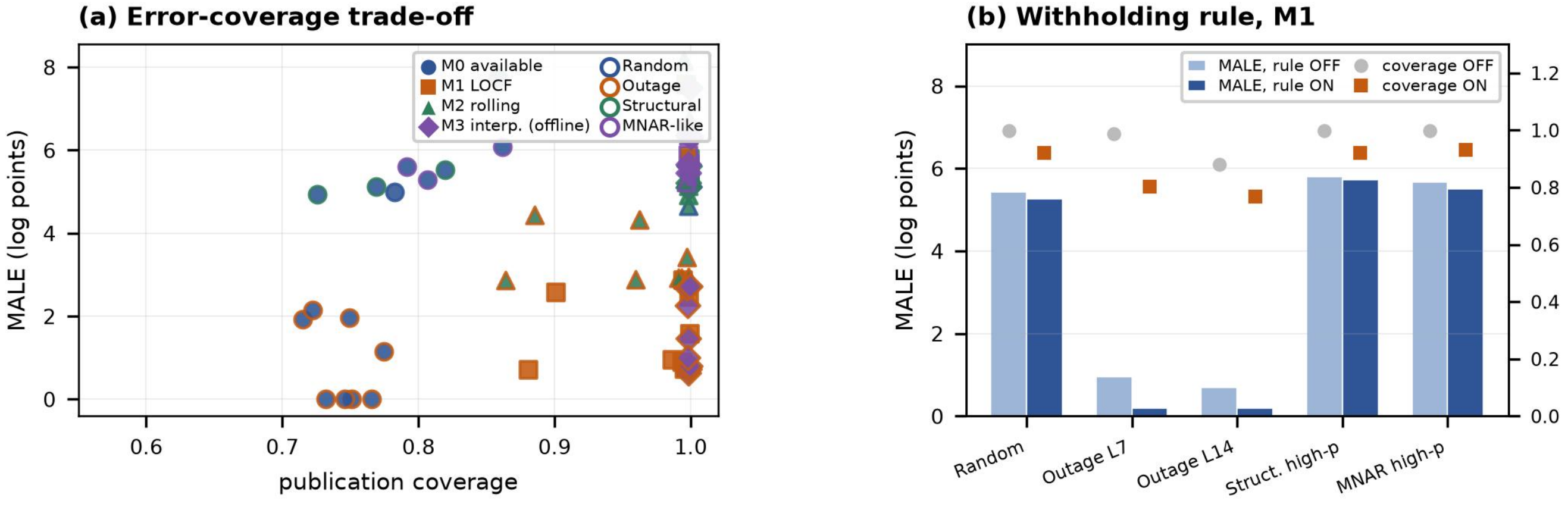

Fig. 3. (a) Conditional error versus publication coverage; (b) M1 withholding at λ = 0.05 (M3 is offline).

### F. Hypothesis assessment

H1–H3 were specified before evaluation on the independent test period.

H1 (mechanism-specific error): descriptively supported. Across 64 nominal-rate comparisons, conditional MALE ranges 0.00–8.14; outages and composition loss differ in availability and level error. Unpaired mechanisms and shared historical data preclude confirmatory inference.

H2 (method ranking): partially supported. Of 16 main-family scenarios, M2 uniquely wins common-date MALE in four; 12 are structural ties. M3 leads adjusted loss in 13 but is offline. Rankings and online-only counts appear in S7 and do not generalise beyond this window.

H3 (coverage-aware withholding): unsupported at λ = 0.05. Although conditional error falls in all 10 cases, adjusted error–coverage performance improves in none; the verdict varies with the failure cost.

### G. Stratum-level reconstruction ablation

Unlike M0–M3, M4 reconstructs individual strata before aggregation. In paired tests (Table VII), M4 lowers MALE in six of seven scenarios and all five composition-changing cases (mean improvement 5.09 log points). Whole-source outages show identical M1/M4 error and coverage because their gate and fallback are shared, irrespective of whether earlier stratum history exists. M4's richer information set limits interpretation.

**TABLE VII. M1–M4 STRATUM-LEVEL ABLATION**

| Scenario | M1 | M4 | Δ | Cov. M1/M4 |
|---|---|---|---|---|
| Random 20% | 5.39 | 1.04 | -4.35 | 1.00 / 1.00 |
| Outage whole 7d | 0.94 | 0.94 | 0.00 | 0.99 / 0.99 |
| Struct. low-price | 7.56 | 0.86 | -6.70 | 1.00 / 1.00 |
| Struct. high-price | 5.77 | 1.09 | -4.68 | 1.00 / 1.00 |
| Struct. 8+ config | 5.45 | 1.30 | -4.15 | 1.00 / 1.00 |
| MNAR-like high-price | 5.67 | 1.08 | -4.59 | 1.00 / 1.00 |
| MNAR-like low-gauge | 6.52 | 1.18 | -5.34 | 1.00 / 1.00 |

Negative Δ favours M4; methods share 1,400 paired masks and coverage. Independent masks explain differences from Table III; details: S5.

### H. Parameter sensitivity

Varying gauge-ratio thresholds (0.50/0.60/0.70), minimum visible cells (2/3/4) and M2 window (3/7/14 days) preserves the qualitative random-versus-whole-outage contrast in 9/9 settings (S4). A separate reference-eligibility check at 2–5 cells preserves it in 4/4 settings (S8). Coverage levels vary, and neither check provides external validation.

## VI. DISCUSSION AND LIMITATIONS

### A. What the mechanism comparison actually shows

The failure mode determines the operator response. Complete-source gaps should be marked unavailable and, if needed, backfilled retrospectively; stratum-composition loss motivates retaining per-stratum histories for past-only reconstruction. Error, coverage and trend fidelity should be reported together rather than selecting a method by conditional MALE alone.

### B. Theoretical connections supported by the results

Imposed missingness mechanisms cannot identify real collection failures [1], [2], [7]. Withholding changes the evaluation sample [3], while online carrying and offline interpolation use different information [4]. Failure penalties express an assumed preference, not observed utility.

Together these findings support a joint measurement-and-availability interpretation.

The gauge-weighted reference-rule check changes M1 MALE from 5.31 to 2.13 (Table VI), illustrating aggregation sensitivity. Under the fixed median, preserving stratum histories enables past-only reconstruction and lowers error by 4.15–6.70 log points in five composition-changing cases (Table VII). This gain reflects finer information and reconstruction; composition-drift alerts remain untested.

### *C. Limitations*

Limitations include one exporter, one 190-day test window, at most six strata per GPU and asking rather than transaction prices. True missingness causes are unidentifiable, and imposed masks are not observed incidents. M4 uses more information than M0–M3; all headline results depend on the equal-weight median. Monte Carlo and bootstrap ranges do not establish cross-market generalisation, and withholding costs are uncalibrated.

## VII. CONCLUSION

This article shows that incomplete GPU-market observations create distinct publication and accuracy failures. Whole-source outages reduce availability; composition-changing masks bias published levels. Reporting conditional error with coverage reveals selection effects, and trend metrics expose smoothing distortion. A separate stratum-level ablation reduces MALE from 7.56 to 0.86 under low-price removal by using richer past information (Table VII). Under the primary failure cost, withholding improves the joint criterion in 0/10 cases, so H3 remains unsupported. Findings concern observed asking-price benchmarks within one historical window.